# MARKERLESS HUMAN MOTION CAPTURE FOR GAIT ANALYSIS

J. Saboune* and F. Charpillet*

*INRIA-LORIA, B.P.239, 54506 Vandœuvre-lès-Nancy, France

saboune@loria.fr, charpillet@loria.fr

**Abstract:** The aim of our study is to detect balance disorders and a tendency towards the falls in the elderly, knowing gait parameters. In this paper we present a new tool for gait analysis based on markerless human motion capture, from camera feeds. The system introduced here, recovers the 3D positions of several key points of the human body while walking. Foreground segmentation, an articulated body model and particle filtering are basic elements of our approach. No dynamic model is used thus this system can be described as generic and simple to implement. A modified particle filtering algorithm, which we call *Interval Particle Filtering*, is used to reorganise and search through the model's configurations search space in a deterministic optimal way. This algorithm was able to perform human movement tracking with success. Results from the treatment of a single cam feeds are shown and compared to results obtained using a marker based human motion capture system.

## Introduction

Referring to several studies, falls are the major cause of accidental mortality in the elderly. This could be considered as a serious problem in societies with growing age average, and thus, reducing the number of falls among seniors is becoming an objective for many research programs. Most of the falls occur during walking and several cross-sectional studies have revealed significant changes in gait patterns associated with advancing age. Many methods have been proposed to detect balance alteration in the elderly while walking. Menz et al. [1] use accelerometers attached to the body in order to evaluate the acceleration patterns at the head and pelvis. Those patterns would then be used to differentiate young and healthy people from old people with risks of falling. The GAITRite system [2] uses a pressure sensors carpet to measure many gait parameters in order to determine dynamic balance and predict fall risk. Depending on the aptitude of a person to accomplish combined physical-psychological tests (Tinetti, Berg, 8foot up and go), geriatricians can detect balance disorders. The purpose of our study is to propose a methodology and a technology to detect a tendency towards the fall of a senior, while observing his daily activities at home. In fact a personal dynamic balance indicator would be evaluated using the gait parameters values. In the case of a weak dynamic balance indication, rehabilitation programs could then be accomplished in order to lower the risk of falling. This approach was set up with the help of experts in the domain of geriatrics and rehabilitation. The originality of our approach generates many constraints to the methods and technologies used. Actually, attaching wearable sensors to the body is prohibited and the senior living environment should not be altered. Our system must be capable of evaluating the balance automatically without any human intervention. On the other hand, used sensors should be low cost. Knowing this, using video feeds from conventional cameras seems to be the most adequate way to measure gait parameters. In order to respect private life, no images should be transferred outside the senior's home and the image processing will be done locally. In this paper we will present a new method to track body 3D motion, which respects the principles of our approach.

## Materials and Methods

### 3D body motion capture

An in-depth gait analysis requires the knowledge of elementary spatio-temporal parameters such as walking speed, hip and knee angles, stride length and width, time of support, among others. In order to obtain this information, a 3D human motion capture system has to be developed. Marker-based systems [3] have been widely used for years with applications found in biometrics. In typical systems, a number of reflective markers are attached to several key points of the patient's body and then captured by infrared cameras fixed at known positions in the footage environment. The markers positions are then transformed into 3D positions using triangulation from the several cameras feeds, making it impossible to track a point's motion when it is not visible by two or more cameras. However, using markers could be considered obtrusive. It also implicates the use of expensive specialized equipment and requires a footage taken in a specially arranged environment. Using video feeds from conventional cameras and without the use of special hardware, implicates the development of a marker less body motion capture system. Research in this domain is generally based on the articulated-models approach. Haritaoglu et al. [4] present an efficient system capable of tracking 2D body motion using a single camera. This might be used in many applications. However it is unable to provide 3D positions, restricting the information we can extract from the feeds. Bregler et al.

[5] used gradient descent search with frame-to-frame region based matching and applied this method on short multi camera sequences. This method proved to be unable to track agile motions with cluttered backgrounds. On the other hand, locating body parts by matching image regions, risks to produce a drift in long sequences. Combining 2D tracker and learned 3D configuration models, Howe et al. [6] were able to produce 3D body pose from short single camera feeds. Gavrila and Davis [7] use an explicit hierarchical search, in which they sequentially locate parts of the body's *kinematic* chain (e.g. torso, followed by upper arm, lower arm and then hand), reducing the search complexity significantly. In real world situations, it seems to be hard to specify each body part in the image independently without using labels or colour cues. Sidenbladh et al. [8] use Condensation algorithm [9] with learned stochastic models and a generative model of image formation to track full body motion. The large number of particles used, makes this algorithm run slowly. Cohen et al. [10] tried to reduce the number of particles, using Support Vector Machine, to train body models. However, using dynamic models would restrict the generality of the approach and prevent the system from tracking gait abnormalities. Using multiple cameras feeds at a 60 frames/s capture, Deutscher et al. [11] produced the best known results to date in 3D full body tracking. Their approach was based on weak dynamical modelling and on *annealed particle filtering*, which is a complex modified Condensation algorithm. In fact, a multi layered particle based stochastic search algorithm was applied to reduce the number of particles. This algorithm uses a continuation principle. Applying this type of layered search augments the risks of falling into local minima, especially in the case of a lower frequency capture, and did not prove to reduce the time complexity in a significant way.

Most of these methods were originally developed and used for character animation and do not meet the requirements of our study. In fact we need to extract the exact 3D position of several points of the human body in order to detect gait abnormalities, using conventional digital camera feeds (25 frames/s) only. Respecting these conditions requires conceiving a new simple algorithm. The method we present here is based on a simple modified particle filtering algorithm, which we call *Interval Particle Filtering (IPF)*. Image foreground segmentation and 3D articulated body modelling are basic elements in our approach.

*The 3D articulated model and likelihood*

The human body will be represented by a 3D articulated model formed by 19 points or joints that represent key elements of the human body (head, elbows, sacrum, knees, ankles etc.).These points are joined up using 17 rigid segments. As in human body, each joint is given a number of degrees of freedom (3 per joint maximum) representing the rotations about the 3D axes (x, y, z). Due to restrictions on human body parts motion we can reduce the number of degrees of freedom assigned to each joint. This model should simulate the human motion. On the other hand, using a large number of degrees of freedom would increase the complexity of the methods used. In fact we opted to use a 31 degrees of freedom model which proved able to simulate approximately the body motion. For each degree of freedom, we define a range beyond which no movement is allowed. These constraints can easily be modified depending on the nature of the actions to be tracked. For example, in a standard walking situation, the leg's rotation cannot take values greater than 60 degrees nor lesser than -30 degrees. Due to the nature of the human body, our model is composed of *kinematic* chains; a body part whose movement implicates the movement of another body part, forms a *kinematic* chain with the latter. Our skeleton model is then fleshed out in a way to have a visually realistic body representation (Figure 1).

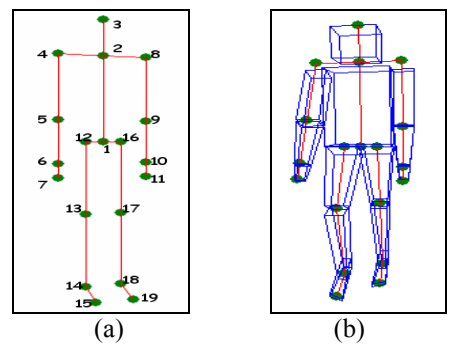

Figure 1: Articulated Model; (a) The skeleton articulated model defined by 19 key points of the human body and 17 segments. This model is composed of 4 *kinematic* chains. Each point is given a number of freedom degrees (3D rotations). (b) The model is fleshed out by adding volumes around the segments.

Deutscher et al. [11] combined a similar model with a dynamic modelling approach; in fact, they use a velocity model for each joint's motion from the previous frame, in order to predict the body pose in the next frame, which restricts the capacity to detect sudden changes in movement if the frames are distant in time. Sidenbladh et al. [8] introduced trained dynamical models. Trained models are of great interest for robust tracking, but they force the real motions tracked to be similar to those observed in the training set. The use of these models would make it unable to detect the abnormal movements we are interested in, due to the fact that these movements would not necessarily be in the training set. In addition it seems impractical to pre-train models for each possible situation and movement of the body. The use of trained models being in contradiction with the goal of our study, we decided not to use any. No dynamic model was used neither, which makes our approach simple and generic.

Knowing the model pose established through its 31 degrees of freedom configuration, we need to find a method to estimate how well this 3D pose fits with the real body pose represented in 2D through the video sequence. The degree of similarity between the real and the estimated pose will be evaluated using a likelihood function. In a particle filtering context the likelihood function is called *weight*. In marker-based systems, the markers positions in each camera's image plan give us the real 3D positions of the markers. In some approaches [8] [10], texture mapping was used to realistically render body images. Despite the advantages it provides, creating this type of images would be specific for each person and the conditions of the video capture (light, clothes etc.). Edge detection and foreground segmentation were used to construct a simple and general likelihood function in [11]. We chose to use a simple foreground segmentation to construct our function. Actually, we construct a *silhouette* image by subtracting pixel by pixel the background from the current image and then applying a threshold filter. This image will then be compared to the synthetic image representing our model's configuration (2D projection of our 3D body's model) to which we want to assign the *weight*. After subtracting the synthetic image from the silhouette, the weight function will be calculated by:

$$w = \frac{N_c}{N_s + N_m}$$

where: $N_c$ = Number of common pixels between the *silhouette* and the synthetic image
$N_s$ = Number of pixels representing the *silhouette* - $N_c$
$N_m$ = Number of pixels of the synthetic image representing the model - $N_c$

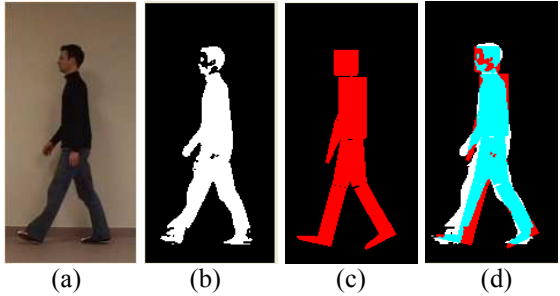

(a)  (b)  (c)  (d)

Figure 2: Estimating the likelihood; The input image (a) will be transformed to a *silhouette* image (b) by subtracting pixel by pixel the background image and then applying a threshold filter. (c) is the synthetic 2D image representing one of the 3D model's configuration. In (d) we compare the synthetic image to the *silhouette* by subtracting the first from the second and evaluating the number $N_c$ of common pixels, the number of pixels of the silhouette outside the synthetic image $N_s$ and the number $N_m$ of pixels of the synthetic image outside the silhouette.

The choice of this function is motivated by the fact that we aim to find the model's configuration that maximises the likelihood of its 2D projection to the real body pose. This can be interpreted by a higher $N_c$ and a lower ($N_s + N_m$). In case of multiple cameras, the weight function $w$ would be:

$$w = \frac{\sum_{i=1}^{c} w_i}{c}$$

where $c$ is the number of cameras, and $w_i$ the weight deducted from the image of camera $i$. This method is simple and can be applied in any condition; however, it may present a little weakness in presence of heavy shadows. This problem can be solved by adjusting the threshold filter.

*Using Interval Particle Filtering*

Full body motion tracking can be treated as a Bayesian state estimation problem. Our 3D model pose is established through the configuration of the degrees of freedom values; This configuration can thus represent the state vector of the model. In addition to the state model, we define an observation**,** through which the likelihood of a state vector at t=$t_k$ is evaluated. In our approach, this observation would be the image of the person we are tracking and the *weight* of a configuration, would represent the observation probability. Deutscher [12] proved that the posterior density in human motion capture is non Gaussian and multi-modal. Particle filtering, also known as the Condensation algorithm [9], proved to be able to handle such type of non-Gaussian, multi-modal densities. In fact it can model uncertainty by transmitting less fitting state configurations at $t_k$, to later time steps, and thus giving them a chance to be chosen. In a particle filtering framework, each 3D model's configuration (31 degrees of freedom vector) is represented by a particle. For each particle a weight is assigned (as described earlier). The particle filtering can be viewed as a search for the best particle in a well defined particles set at each time step.

In order to have a realistic state vector estimation, a certain number of particles are necessary. In a high dimensional space this number becomes relatively big. The use of a greater number of particles leads to better results. On the other hand, using more particles augments the temporal complexity of the algorithm, due to the fact that at each time step the *weight* for all particles must be calculated. The goal of all modified particle filtering algorithms is to reduce the number of particles needed, and this especially in high dimensional spaces, where the complexity could make the basic particle filtering algorithm practically inapplicable. Despite the high dimension of our state vector (31 degrees of freedom), we opted to use a particle filtering algorithm, due to its capacity to handle the multi modal observation probability. In parallel, we modified the basic algorithm by introducing the *Interval Particle Filtering* that tends to reconfigure the particles search space in an optimal way.

The *Interval Particle Filtering (IPF)* introduces simple modifications on the Condensation algorithm in order to optimise the particles search space configuration. These modifications are done in a way to preserve the advantages a particle filter offers. Neither dynamic modelling nor an evolution model was used. A single iteration per time step is accomplished, excluding any layered search. We preserve the 3 steps structure of the basic particle filtering algorithm at each time step $t_k$ :

- *Selection*: The N particles set created at $t_{k-1}$ is sorted by its weights. In this sorted set, a number of *M* distinct particles, that have the biggest *weights*, are selected.
- *Prediction*: As we are especially interested in some specific elements of the state vector, we can dissociate it into 2 state vectors: the first one **L** containing the 'interesting' degrees of freedom and the second **R** covering the rest of the freedom degrees. Each particle is now represented by 2 vectors instead of one. **L** is then updated and replaced by a multidimensional Interval (space) composed of *I* vectors covering a grid of vectors surrounding the initial vector **L**. This approach is inspired by the presence of physiological restrictions on the degrees of freedom evolution (e.g. maximum angular velocity of joints in human motion). As a result, each particle will be updated and replaced by *I* 'neighbour' particles so as to cover the whole possible configuration space of the 'interesting' dimensions, closely surrounding the particle's 'interesting' dimensions configuration. For each particle, **R** is then updated by adding a white noise vector. The width of the interval and the number *I* of vectors depend on the nature of the system. A wider interval and a greater *I* provide more accurate results but result in greater computational cost.
- *Measure*: This step remains unaltered; given the observation (image at $t_k$), the *weight* for each particle is calculated and the new *weighed* particles set is propagated to be used at $t_{k+1}$. The estimated state vector (body pose) at $t_k$ will be represented by the particle (the model's configuration) having the greatest *weight*.

In presence of restrictions on the state vector evolution, *Interval Particle Filtering* reorganises the set of N particles into *M* sub-sets each formed of *I* particles covering in a deterministic way the 'neighbourhood' of the *j* heaviest particle *(j=1..M)* at the previous time step.

### Results

We use video feeds captured at 20 Hz, and due to the physiological constraints, we can define the interval width for each angle to be 10°. This means that the angle between two time steps (50 ms) can not evolve (positively or negatively) for more than 5°. If each interval contains 81 vectors and *M*=81 our algorithm will be running with 6561 particles. At each time step we can get the estimated 3D positions of the 19 points forming the articulated model and the estimated values of the 31 freedom degrees. The initialisation of the body parts configuration is done automatically. In fact an exhaustive search is applied to the initial set of particles in order to find the *heaviest* particle. This initial set contains N vectors configured as to cover a well defined grid of plausible configurations.

We present results from video captures of three different subjects, moving in a normal environment. Those feeds were captured at 20 frames/s using a single commercial digital video camera. Image processing was done offline, using a P4 3GHz PC. It takes 20 seconds per frame (7 minutes per second of footage) to find the body origin 3D position and its parts configuration using *Interval Particle Filtering*. Despite being too far from a real time performance, our system runs faster than many other systems developed in the literature. The first set of images (Figure 3) shows a person walking in a straight line, with momentary occlusions.

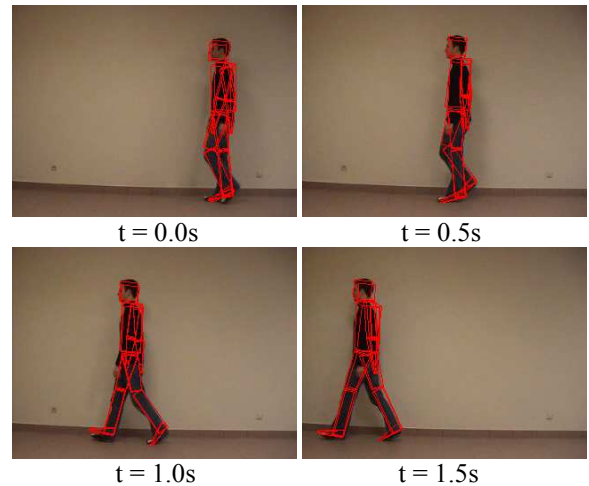

| t = 0.0s | t = 0.5s |
| t = 1.0s | t = 1.5s |

Figure 3: Results for a subject walking normally along a wall; despite the fact that the left leg is occluded in some frames, the algorithm was able to track the movement with success.

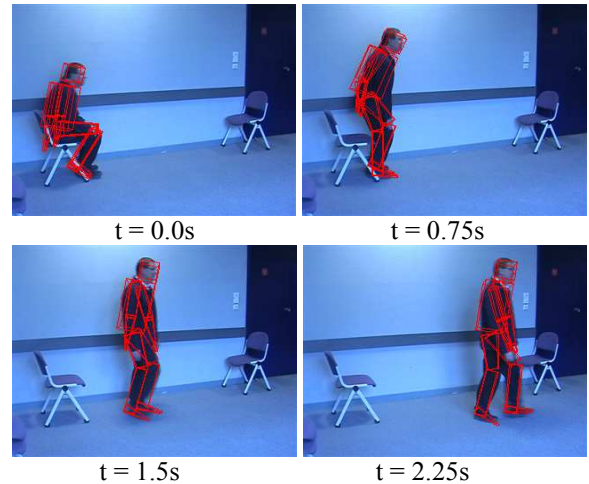

| t = 0.0s | t = 0.75s |
| t = 1.5s | t = 2.25s |

Figure 4: Images showing a subject getting up from a chair, turning and then walking. This set of images illustrates the use of the body origin positioning algorithm we developed. The challenge in this feed is to determine the body origin position for the person while getting up and beginning to walk.

The second set (Figure 4) shows a person getting up from a chair and then turning and walking. The difficulty here is to track the movement of the person while getting up, but the algorithm succeeded in it. The third set (Figure 5) shows a female subject walking in a random manner. The challenge here is that the legs are occluded by the subject's skirt. However due to the use of *IPF* this difficulty has been surmounted. The last set (Figure 6) introduces a person walking randomly. This scene had also been filmed and treated by a Vicon system (marker based motion capture system) running at 100 Hz. Positions of some body points produced by our algorithm can be favourably compared to the same positions produced by the Vicon, as shown in Figures 7.

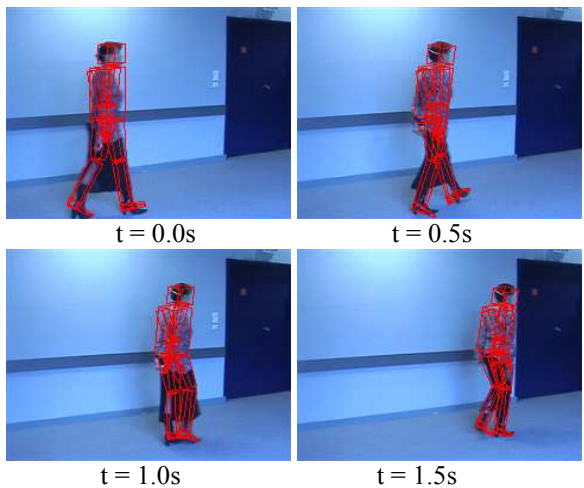

| t = 0.0s | t = 0.5s |
| t = 1.0s | t = 1.5s |

Figure 5: Images showing a female subject wearing a long skirt walking randomly. Despite the fact that the legs were occluded in these images, our system is capable of recovering and estimating their positions.

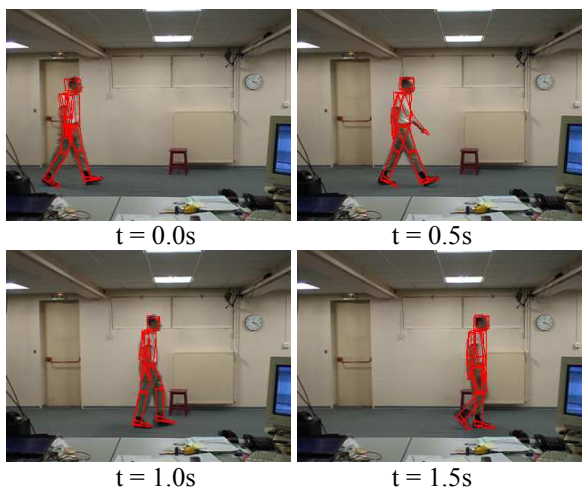

| t = 0.0s | t = 0.5s |
| t = 1.0s | t = 1.5s |

Figure 6: Sequence showing a subject walking randomly. This scene had been also captured by a Vicon system composed of 6 infra red cameras. Reflective markers were fixed to the subject's body key points (the same key points of our skeleton model) and the results were compared to the results obtained using IPF algorithm (Figures 7).

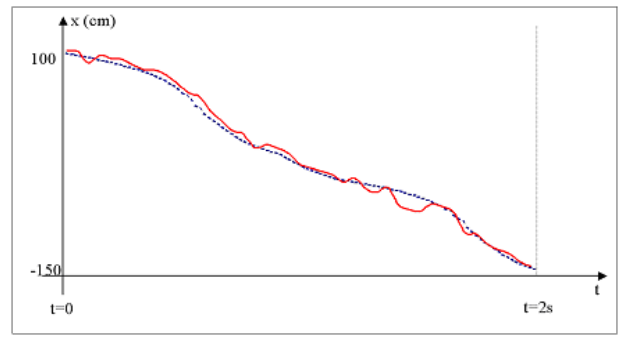

Figure 7.1: Longitudinal displacement x (calculated in the camera referential) of the right knee (for the subject appearing in Figure 6), estimated by IPF (in plain) and using Vicon system (in dotted). The two curves have similar shapes but the dotted curve is smoother. This can be explained by the differences in capture and treatment frequency (20 Hz for IPF, 100 Hz for Vicon).

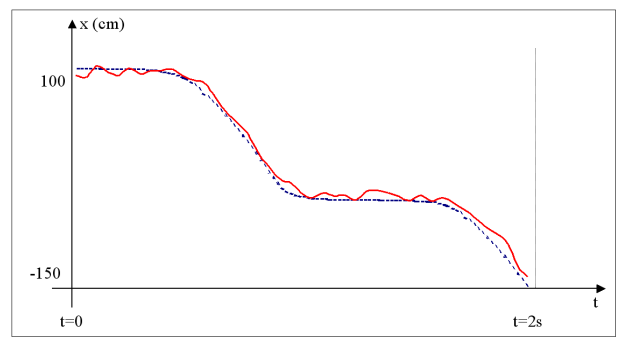

Figure 7.2: Longitudinal displacement x (calculated in the camera referential) of the right ankle (for the subject appearing in Figure 6), estimated by IPF (in plain) and using Vicon system (in dotted).

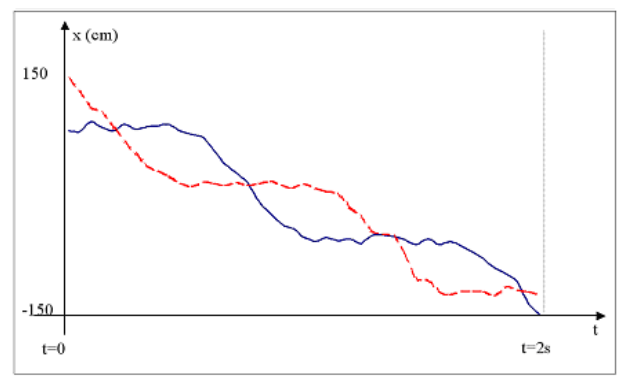

Figure 8: Longitudinal displacement x (calculated in the camera referential) of the right ankle (in plain) and the left ankle (in dotted) estimated both by IPF. Many parameters could be extracted from this graphic, such as time of double support, step length etc..

**Discussion**

The fact of using a single camera prevents us from seeing some body parts (which depends on the view angle) for a long portion of the footage and the movement of these parts could thus not be evaluated fairly. Using multiple cameras would solve this problem but implicates the complexity of calibrating and adjusting a stereo vision system. The system developed will be tested to analyse the gait of senior fallers during an experiment which will be held at a geriatrics department of a hospital. In fact the senior will be asked to follow a specially arranged path and the gait parameters would be evaluated and analysed in order to evaluate the pertinence of some variables.

**Conclusion**

In this paper we presented a new approach for marker less human motion capture from a single commercial video camera, based on a modified particle filtering algorithm. The aim of our study is to extract gait parameters from the feeds, and the results obtained compared to those of a marker based system were encouraging. The *Interval Particle Filtering* we introduced here, proved to give good results despite the high dimensionality of the state vector, even with occlusions. This algorithm is simple to implement and works with video feeds captured at any frequency and in any environment. Since we do not use any restrictive dynamic model, our approach can be also described as being generic.

**Acknowledgements**

This work was financed by the RNTS project *PARAChute* and the INRIA action *DialHemo*.